\documentclass{midl} % Include author names
\usepackage{mwe} % to get dummy images
\usepackage{graphicx}
\usepackage{subcaption}
\usepackage{multirow}
\jmlrvolume{-- Under Review}
\jmlryear{2019}
\jmlrworkshop{Full Paper -- MIDL 2019 submission}
\editors{Under Review for MIDL 2019}

\title[]{Joint shape learning and segmentation for medical images using a minimalistic deep network}

\midlauthor{\Name{Balamurali Murugesan \nametag{$^{1,2}$}} \Email{balamurali@htic.iitm.ac.in}\\
\addr $^{1}$ Indian Institute of Technology Madras (IITM), India \\
\addr $^{2}$ Healthcare Technology Innovation Centre (HTIC), IITM, India \AND
\Name{Kaushik Sarveswaran \midljointauthortext{Work was done while interning at HTIC.} \nametag{$^{2,3}$}} \Email{ced14i032@iiitdm.ac.in}\\
\addr $^{3}$ Indian Institute of Information Technology \\ Design \& Manufacturing Kancheepuram (IIITDM), India\\
\Name{Sharath M Shankaranarayana \nametag{$^{1,4}$}} \Email{sharath@zasti.ai} \\
\addr $^{4}$ Zasti, India \\
\Name{Keerthi Ram \nametag{$^{2}$}} \Email{keerthi@htic.iitm.ac.in} \\
\Name{Mohanasankar Sivaprakasam \nametag{$^{1,2}$}} \Email{mohan@ee.iitm.ac.in} \\
}
\begin{document}

% balamurali@htic.iitm.ac.in,sharath@zasti.ai,ced14i032@iiitdm.ac.in,keerthi@htic.iitm.ac.in,mohan@ee.iitm.ac.in
\maketitle

\begin{abstract}
Recently, state-of-the-art results have been achieved in semantic segmentation using fully convolutional networks (FCNs). Most of these networks employ encoder-decoder style architecture similar to U-Net and are trained with images and the corresponding segmentation maps as a pixel-wise classification task. Such frameworks only exploit class information by using the ground truth segmentation maps. In this paper, we propose a multi-task learning framework with the main aim of exploiting structural and spatial information along with the class information. We modify the decoder part of the FCN to exploit class information and the structural information as well. We intend to do this while also keeping the parameters of the network as low as possible. We obtain the structural information using either of the two ways- i) using the contour map and ii) using the distance map, both of which can be obtained from ground truth segmentation maps with no additional annotation costs. We also explore different ways in which distance maps can be computed and study the effects of different distance maps on the segmentation performance. We also experiment extensively on two different medical image segmentation applications- i.e i) using color fundus images for optic disc and cup segmentation and ii) using endoscopic images for polyp segmentation. Through our experiments, we report results comparable to, and in some cases performing better than the current state-of-the-art architectures and with an order of 2x reduction in the number of parameters. 
%Our proposed scheme has an advantage that improved segmentation results can be obtained without complicated post-processing techniques. 

\end{abstract}


\section{Introduction}
% Semantic segmentation is an important task in many of medical imaging problems. Fully convolutional networks (FCNs) such as U-net \cite{U-Net} have been shown to be highly suitable for the task of semantic segmentation of medical images for almost all modalities \cite{ker2018deep}. 
Image segmentation is the process of delineating regions of interest in an image, which is one of the primary tasks in medical imaging. Identifying these regions provides numerous applications in the medical domain. Some of the applications include: segmentation of glands in histology images which is an indicator of cancer severity, segmentation of optic cup and disc in retinal fundus images which is used in glaucoma screening, segmentation of lung nodules in chest computed tomography which aids physicians in differentiating malignant lesions from benign lesions, and segmentation of polyp in colonoscopy images which helps in diagnosing cancer in its early stages \cite{survey}. 
%Something about difficulty The difficulty of medical image segmentation is due to their variability. 

Traditional approaches in image segmentation include active contours \cite{active_contour}, normalized cuts \cite{normalized_cut} and random walk \cite{random_walk}. Recently, fully convolutional networks (FCNs) such as U-Net \cite{U-Net} have been shown to be highly suitable for the task of semantic segmentation of medical images for almost all modalities \cite{ker2018deep} and have been shown to achieve better results than the traditional methods. The current trend in the research using U-Net mainly revolves around two approaches: $i)$ the first approach focuses on modifying the U-Net architecture by adding residual, dense and multi-scale blocks \cite{intro_residual, resunet}, $ii)$ the second approach focuses on modifying the loss functions by adding dice, jaccard coefficients to normal cross-entropy while fixing U-Net as the base network \cite{intro_dice}. These approaches have provided a significant improvement in segmentation. However, medical images require segmentations to be of greater precision. One of the reasons for segmentation to be of poor precision is that the region of interests usually occupy a small area in medical images, resulting in severe foreground-background class imbalance, which leads to imprecise segmentation. Also, the network lacks the knowledge of an object's shape because of spatial information being lost in encoder through max-pooling which results in irregular segmentation. Recently multiple works \cite{intro_skin,intro_shape}, have addressed these issues. But such networks are not capable of explicitly learning the spatial information. To this end, we propose a novel architecture which is capable of learning both class and spatial information explicitly through a joint learning framework. There are two related works which are of our interest: 1) The network DCAN proposed by \cite{dcan} provided better segmentation results with help of contours. 2) The network proposed by \cite{isbi_dcan} showed improvement over DCAN with the help of distance maps. Both these methods propose the use of FCNs with the architecture having single encoder block and two decoder blocks, where one of the decoders is dedicated for segmentation task and the other is dedicated for the auxiliary task. But the networks proposed have complicated architectures and have large number of parameters, leading to longer training time and inference time, while also requiring more compute resources.  
\begin{figure}
    \centering
    \includegraphics[width=15cm]{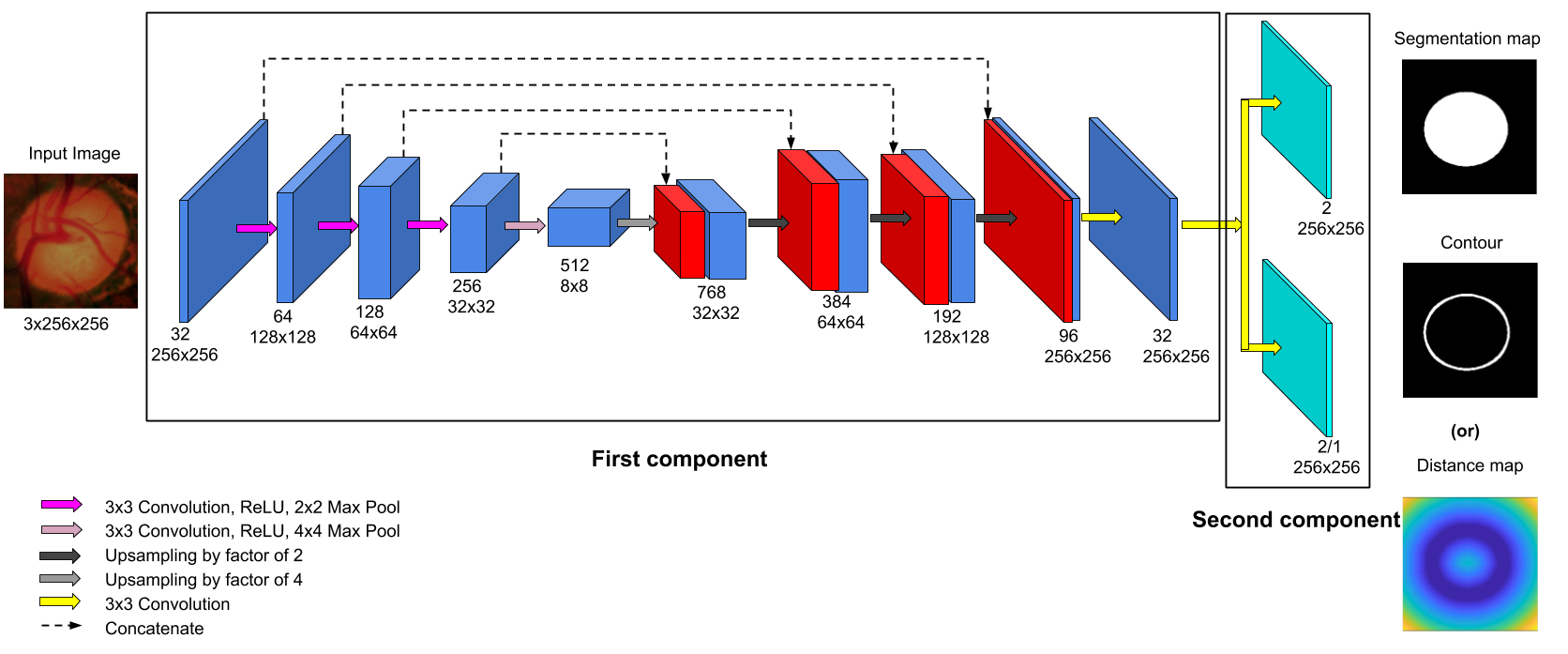}
    \caption{Network Architecture with U-Net architecture followed by two parallel convolution blocks}
    \label{fig:arch}
\end{figure} 
In this paper, as the main contribution, we propose a minimalistic deep network for the task of joint shape learning and segmentation. The proposed architecture consists of significantly fewer parameters while maintaining the performance and in many cases even outperforming the previous methods. We also explore numerous ways in which the spatial information can be incorporated and study their effects on the performance. We conduct multiple experiments for two different kinds of medical images- for optic disc and cup segmentation from retinal color fundus images and polyp segmentation from endoscopic images and report state of the art results. 

% The paper is organized as follows: Section \ref{sect:architecture} introduces the proposed architecture, Section \ref{sect:loss_function} intro

\section{Methodology}
In this section, we first present the novel end-to-end multi-task architecture for improving semantic segmentation, which is capable of exploiting spatial and structural information along with the class information, while also keeping the number of parameters less. We then present the ways in which structural information was obtained to aid the network. Next, we explain how the network was trained to learn the class information and the spatial information using different loss functions.  

\subsection{Network Architecture}\label{sect:architecture}
The proposed architecture is shown in Figure \ref{fig:arch}. The architecture is an FCN consisting of two components. The first component of the network is similar to U-Net \cite{U-Net}. The second component consists of parallel convolutional blocks for multi-task learning. %performing multiple tasks. 
%encoder,decoder? Repeated->doubles=each step? 2 feature maps or no.of classes?

The first component has an encoder-decoder architecture with encoder providing a contracting path and decoder providing an expansive path. The encoder consists of repeated applications of convolutions with kernel size of 3x3 and stride 1, followed by a rectified linear unit (ReLU) activation and 2x2 max pooling with stride 2 for downsampling. Repeated application of filters doubles the number of feature map and halves the feature dimension at each step. The final convolution in the encoder is carried out with 4x4 max pooling with other elements remaining the same. In the decoder path, upsampling is done to feature map by an initial factor of 4, followed by repeated upsampling by a factor of 2. Each feature map in the decoder is concatenated with the corresponding feature map from the encoder. This concatenation helps in retaining the feature maps from different scales. 

As shown in Figure \ref{fig:arch}, the top path in second component is the classification branch responsible for estimating the segmentation mask while the bottom path is used for the auxiliary task and is to estimate either the contour map as a classification task or the distance map as a regression task. For mask and contour estimation, 3x3 convolution is applied to get 2 feature maps while for distance map, the same convolution is applied to get 1 feature map.

% The top path in second component is for estimating mask while the bottom path is used to estimate either the contour or distance map. For mask and contour estimation, 32x3x3 convolution is applied to get 2 feature maps while for distance map, the same convolution is applied to get 1 feature map. 

\begin{figure}
\centering     %%% not \center
\includegraphics[width=50mm]{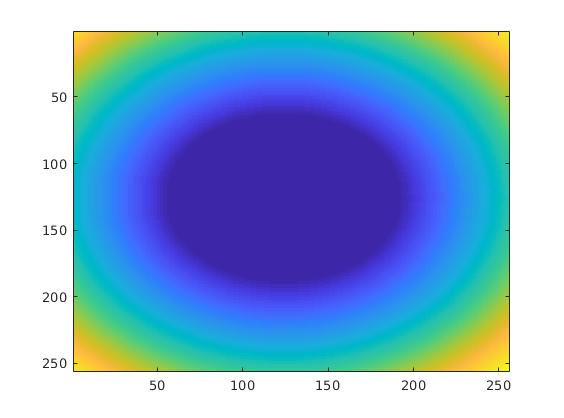}
\includegraphics[width=50mm]{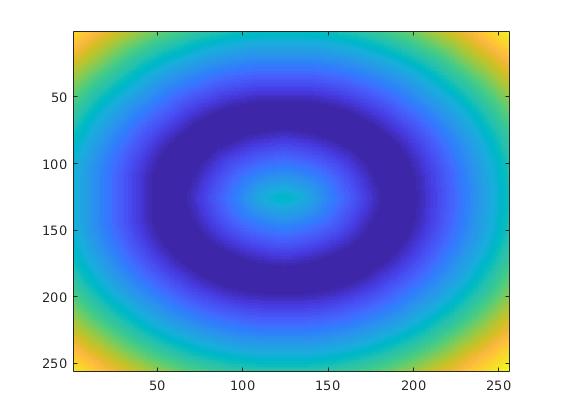}
\includegraphics[width=50mm]{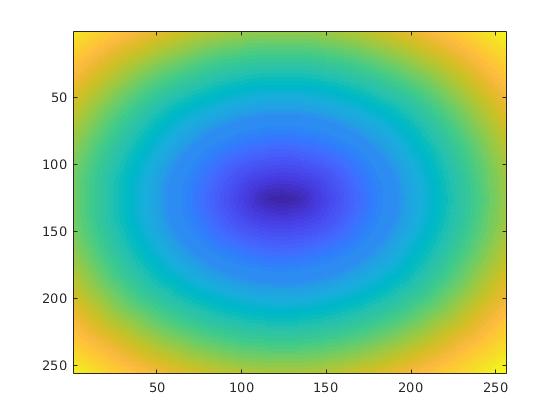}
\includegraphics[width=50mm]{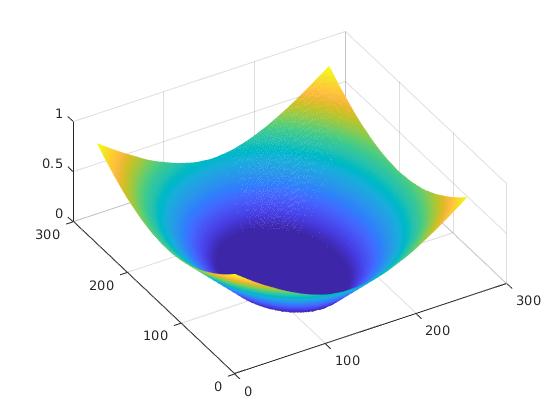}
\includegraphics[width=50mm]{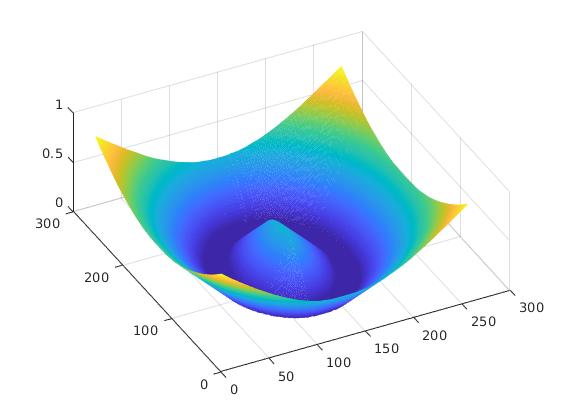}
\includegraphics[width=50mm]{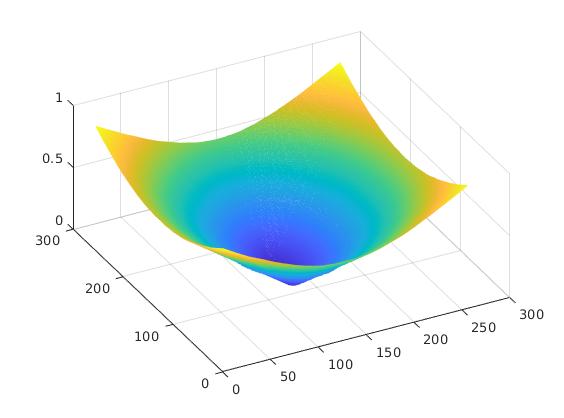}
\caption{Left to right: Distance maps D1, D2, D3}
\label{fig:dist_maps}
\end{figure}

\subsection{Capturing Structural Information}\label{sect:structure_information}
We explored multiple techniques in order to capture spatial and structural information. We harness the spatial information that is implicitly present in ground truth segmentation masks and we achieve the same in two ways: $i)$ using the contours obtained from the segmentation map and $ii)$ using the euclidean distance transforms computed from the segmentation maps. 
For obtaining the contour map C, we first extract the boundaries of connected components based on the ground truth segmentation maps which are subsequently dilated using a disk filter of radius $5$. We also explore various kinds of distance transform maps. Using distance map allows us to assign a value to each pixel in an image relative to the nearest boundary of segmentation map. This alleviates the pixel-wise class imbalances which arise in the segmentation maps. Thus for an image, we assign values $ (d_1, d_2, ... , d_n)$ to all the pixels with $n$ being the total number of pixels and $d_i$ being the distance of the $i_{th}$ pixel to the closest boundary of the mask. 
We propose to use three kinds of distance maps based on the values assigned to the pixels. $i)$ for the first case $D_1$, we assign positive distances for all the points outside the boundary and assign zero values for all the points inside the boundary or the mask region. $ii)$ for the second case $D_2$, we assign positive distances for the points inside and outside boundary while having zero values for the pixels on the boundary. $iii)$ for the third case $D_3$, we assign positive distances for the points outside and negative distances for the points inside the boundary and zero values for the points on the boundary. Figure \ref{fig:dist_maps} is a visualization of the distance maps in 2D and 3D form. We show that the choice of distance map is also an important factor which affects the model performance. 

%Thus, the aforementioned steps help in capturing the spatial and structural information, which are not sufficiently captured in the segmentation maps and also without any additional cost regarding annotations. 

\subsection{Loss function}\label{sect:loss_function}
%The proposed network provides two output maps for a single input image. The network can be split into two cases. In the first case, the network outputs mask, contour pair while in the second case it outputs mask, distance pair. The mask and contour estimation are treated as a classification problem whereas the distance estimation is treated as a regression problem. The most common Negative Log Likelihood is used for classification and Mean Square error is used for classification. The intended mask output is constrained by either contour or distance. The two loss functions for the two possible networks are explained below. 9 
The mask prediction is a classification task and Negative Log Likelihood (NLL) is used as a loss function. The mask prediction is regularized either by contour or distance map learning tasks. For the classification task of contour prediction, NLL is used as a loss function. For the regression task of distance map prediction, Mean Square Eror (MSE) is used as a loss function. The combined loss functions involving the mask-contour pair and the mask-distance pair are formulated below. 

\subsubsection{Contour constraint:}\label{subsect:contour}
The loss term $\mathcal{L}_{mc}$ for using contour map as a constraint is given by 
\begin{equation}
\mathcal{L}_{mc} = \mathcal{L}_{mask} + \mathcal{L}_{contour} 
\end{equation}
where 
\begin{equation}\label{eq:l_mask}
\mathcal{L}_{mask} = \sum_{\boldsymbol{x} \, \epsilon\, \Omega}log\, p_{mask}(\boldsymbol{x};l_{mask}(\boldsymbol{x}))
\end{equation}
\begin{equation}
\mathcal{L}_{contour} = \sum_{\boldsymbol{x}\, \epsilon\, \Omega}log\, p_{contour}(\boldsymbol{x};l_{contour}(\boldsymbol{x})) 
\end{equation}
$\mathcal{L}_{mask}$ and $\mathcal{L}_{contour}$ denotes the pixel-wise classification error. $\boldsymbol{x}$ is the pixel position in image space $\Omega$. $p_{mask}(\boldsymbol{x};l_{mask})$ denotes the predicted probability for true label $l_{mask}$ after softmax activation function. $p_{contour}(\boldsymbol{x};l_{contour})$ denotes the predicted probability for true label $l_{contour}$ after softmax activation function.
\subsubsection{Distance constraint:}\label{subsect:distance}
The loss term $\mathcal{L}_{md}$ for using distance map as a constraint is give by 
\begin{equation}
\mathcal{L}_{md} = \mathcal{L}_{mask} + \mathcal{L}_{distance}
\end{equation}
where 
\begin{equation}
\mathcal{L}_{distance} = \sum_{\boldsymbol{x}\, \epsilon\, \Omega} (\hat{D}(\boldsymbol{x}) - D(\boldsymbol{x}))^2
\end{equation}
$\mathcal{L}_{mask}$ is from equation \ref{eq:l_mask} and $\mathcal{L}_{distance}$ denotes the pixel-wise mean square error. $\hat{D}(\boldsymbol{x})$ is the estimated distance map after sigmoid activation function while $D(\boldsymbol{x})$ is the ground-truth.

% % \mathcal{L}_{contour}
% % \mathcal{L}_{distance}

\section{Experiments and Results}\label{sect:Experiments}

\subsection{Dataset and Pre-processing}\label{sect:Dataset}
We use ORIGA dataset \cite{origa} for the task of optic disc and cup segmentation. The dataset contains $650$ color fundus images along with the pixelwise annotations for the optic disc and the cup. 
We obtain the final segmentation map by thresholding the output probabilities similar to the work in \cite{PT2018}. We then fit an ellipse on the segmentation outputs for both cup and disc. 
\\
\\    
We also use Polyp segmentation dataset from MICCAI 2018 Gastrointestinal Image ANalysis (GIANA) \cite{polyp_dataset} for evaluating the models because polyp has large variations in terms of shape. The dataset consists of 912 images with ground truth masks. The dataset is randomized and split into 70\% for training and 30\% for testing. The images are center-cropped to square dimensions and resized to 256$\times$256 before usage.

\subsection{Implementation Details}\label{sect:implementation}
The models are implemented in PyTorch \cite{pytorch}. Each model is trained for 150 epochs with Adam optimizer, with a learning rate to 1e-4 and batch size of 4. All experimentations are conducted with NVIDIA GeForce GTX 1060 with 6GB vRAM.

\subsection{Results and Discussion} \label{sect:results}%\subsection{Evaluation and Comparison}
The metrics used for evaluating the performance of the network include Jaccard and Dice. The explanation of Jaccard and Dice can be found in Appendix \ref{appendix:A}. Some denotations used in this section are Encoder (Enc), Decoder (Dec), Mask (M), Contour (C), Distance (D) and Parallel convolution block after U-Net (Conv). 
The results of the proposed networks (1Enc 1Dec + Conv MC and 1Enc 1Dec + Conv MD) are compared with the following combinations of networks and loss functions. %These networks are tested for both cup and disc segmentation. 
\begin{itemize}
    \item {A network (1Enc 1Dec M)  \cite{U-Net} with a single encoder and a decoder having NLL as loss function for mask estimation.}
    \item {A network (1Enc 2Dec MC) \cite{dcan} with a single encoder and two decoders having NLL as loss function for both mask and contour estimation.}
    \item {A network (1Enc 2Dec MD) \cite{isbi_dcan} with a single encoder and two decoders having NLL as loss function for mask and MSE as loss function for distance estimation.}
\end{itemize}

\begin{table}[]
\centering
\caption{Comparison of segmentation metrics on different networks and loss functions}
\label{table:metrics1}
\small
\begin{tabular}{|c|c|c|c|c|c|c|}
\hline
\multirow{2}{*}{Architecture} & \multicolumn{2}{c|}{Cup} & \multicolumn{2}{c|}{Disc} & \multicolumn{2}{c|}{Polyp} \\ \cline{2-7} 
 & Dice & Jaccard & Dice & Jaccard & Dice & Jaccard \\ \hline
1Enc 1Dec M \cite{U-Net} & 0.8655 & 0.7712 & 0.9586 & 0.9215 & 0.8125 & 0.7323 \\ \hline
1Enc 2Dec MC \cite{dcan} & 0.8715 & 0.7803 & 0.9646 & 0.9324 & 0.8151 & 0.7391  \\ \hline
1Enc 2Dec MD \cite{isbi_dcan} & 0.8723 & 0.7807 & 0.9665 & 0.9358 & 0.8283 & 0.7482 \\ \hline
1Enc 1Dec + Conv MC (Ours) & 0.8717 & 0.7798 & 0.9643 & 0.9318 & 0.8152 & 0.7383 \\ \hline
1Enc 1Dec + Conv MD (Ours) & 0.8721 & 0.7805 & 0.9662 & 0.9348 & 0.8291 & 0.7514 \\ \hline
\end{tabular}
\end{table}

% Feature maps
\begin{figure}
\centering     %%% not \center
\includegraphics[width=30mm]{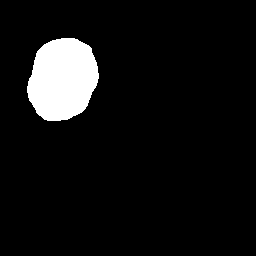}
\hspace{0.5mm}
\includegraphics[width=30mm]{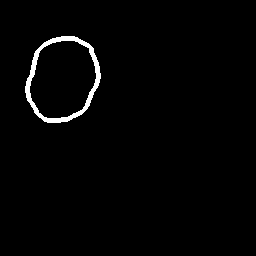}
\hspace{0.5mm}
\includegraphics[width=30mm]{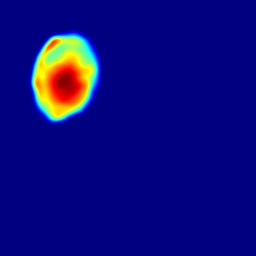}
\hspace{0.5mm}
\includegraphics[width=30mm]{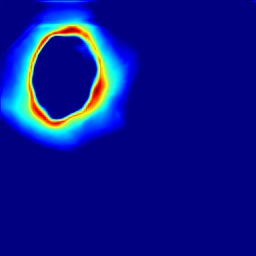}
\vspace{2mm}

\caption{Left to Right: Mask, Contour, Feature map 1 (represents mask), Feature map 16 (represents contour)}
\label{fig:feature_maps}
\end{figure}

\begin{table}[]
\centering
\caption{Comparison of no.of parameters and running time for different networks}
\label{table:metrics2}
\begin{tabular}{|c|c|c|}
\hline
Architecture & Running time (ms) & No. of parameters \\ \hline
1Enc 1Dec M \cite{U-Net} & 1.3131 & 7844256 \\ \hline
1Enc 2Dec MC \cite{dcan} & 1.8677 & 10978272 \\ \hline
1Enc 2Dec MD \cite{isbi_dcan} & 1.8531 & 10977984 \\ \hline
1Enc 1Dec + Conv MC (Ours) & 1.3384 & 7844832 \\ \hline
1Enc 1Dec + Conv MD (Ours) & 1.3235 & 7844544 \\ \hline
\end{tabular}
\end{table}

From Table \ref{table:metrics1} it can be seen that the network 1Enc 2Dec MC and 1Enc 1Dec + Conv MC have similar results for cup, disc and polyp segmentation. Likewise, the network 1Enc 2Dec MD and 1Enc 1Dec + Conv MD have nearly equal results for the cup, disc and polyp segmentation. This shows that having a single decoder with two convolution paths achieves equivalent results to the network with two parallel decoders. This observation indicates that a single decoder in itself is sufficient to reconstruct features of mask, contour and distance from the encoder representation. This reasoning can be supported by visualizing the 32 decoder feature maps obtained before parallel convolution blocks. From Figure \ref{fig:feature_maps}, it can be seen that feature maps 1 and 16 represent the approximation of mask and contour respectively. Similarly, the distance maps can also be obtained through a linear combination of feature maps. The remaining feature maps can be viewed in Appendix \ref{appendix:B}. This validates our claim that only a few convolutions are required post the final decoder layer for obtaining the contour as well as distance maps.    

Also, because of using a single decoder network, the number of parameters involved reduces by half when compared to the networks with two decoders. The addition of parallel convolution path at the end of the decoder has very less effect on the number of parameters.  From Table \ref{table:metrics2}, it can be seen that our proposed networks 1Enc 1Dec + Conv MC and 1Enc 1Dec + Conv MD have a nearly equal number of parameters to 1Enc 1Dec M (U-Net).  And it is also clear that our proposed networks 1Enc 1Dec + Conv MC and 1Enc 1Dec + Conv MD have a 50\% reduction in the number of parameters compared to 1Enc 2Dec MC and 1Enc 2Dec MD. 

The running time is the average time taken by the network to process a single image. The running time of the network depends on the number of parameters. The network with a higher number of parameters will have larger running time compared to the network with less number of parameters. From Table \ref{table:metrics2}, it can be seen that our proposed networks 1Enc 1Dec + Conv MC and 1Enc 1Dec + Conv MD have running time nearly equal to 1Enc 1Dec M (U-Net). And it is also clear that our proposed networks 1Enc 1Dec + Conv MC and 1Enc 1Dec + Conv MD show nearly 1.4$\times$ speed-up compared to 1Enc 2Dec MC and 1Enc 2Dec MD. %The same when increasing the architecture like res, dense this will be crucial.

%sample results
\begin{figure}
\centering     %%% not \center
\hspace{0.5mm}
\includegraphics[width=30mm]{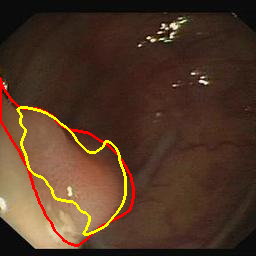}
\hspace{0.5mm}
\includegraphics[width=30mm]{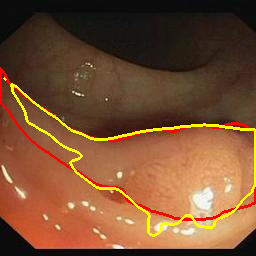}
\hspace{0.5mm}
\includegraphics[width=30mm]{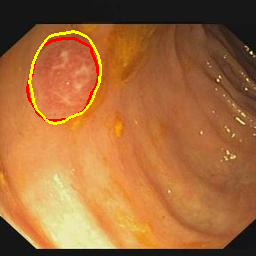}
\hspace{0.5mm}
\includegraphics[width=30mm]{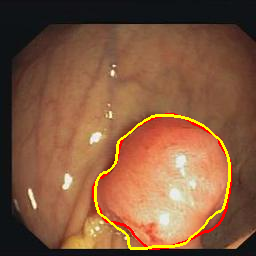}
\vspace{2mm}

\includegraphics[width=30mm]{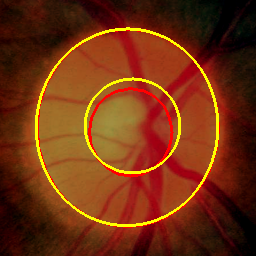}
\hspace{0.5mm}
\includegraphics[width=30mm]{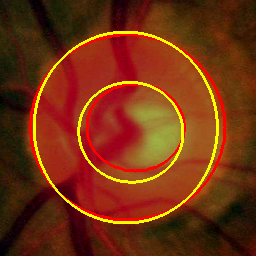}
\hspace{0.5mm}
\includegraphics[width=30mm]{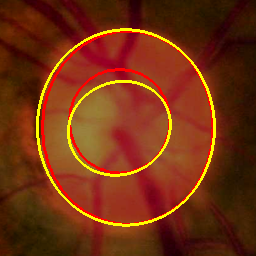}
\hspace{0.5mm}
\includegraphics[width=30mm]{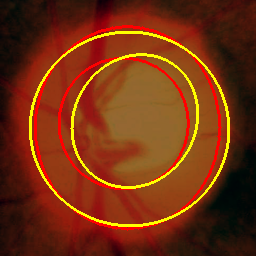}
\vspace{2mm}

\caption{Sample results, Red - Ground truth, Yellow - Predicted}
\label{fig:results1}
\end{figure}

Some of the results obtained using our best network (1Enc 1Dec + Conv MD) are shown in Figure \ref{fig:results1}. In the figure, first row depicts the segmentation results obtained using polyp test data while second row depicts the segmentation results obtained using cup and disc test data. In the images, contour drawn by red color denotes ground truth and contour drawn by yellow color denotes the predicted output.  %If you want o tell about outputs. 

\begin{figure}
\centering     %%% not \center
\includegraphics[width=28mm]{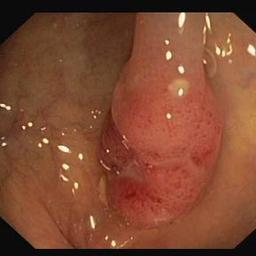}
\hspace{0.5mm}
\includegraphics[width=28mm]{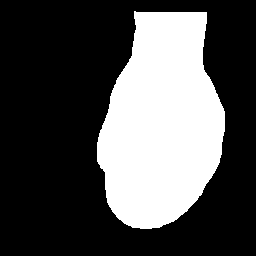}
\hspace{0.5mm}
\includegraphics[width=28mm]{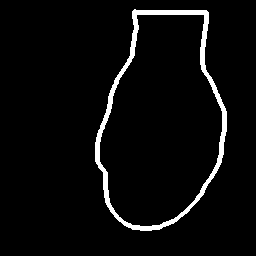}
\hspace{0.5mm}
\includegraphics[width=28mm]{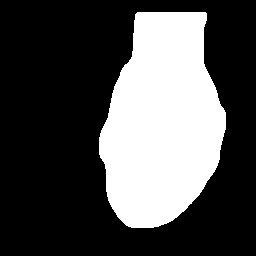}
\hspace{0.5mm}
\includegraphics[width=28mm]{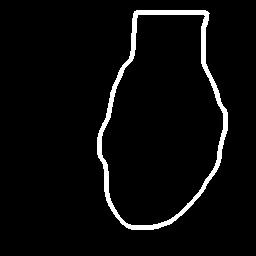}
\vspace{2mm}

\includegraphics[width=28mm]{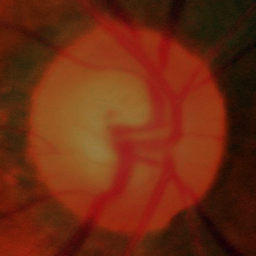}
\hspace{0.5mm}
\includegraphics[width=28mm]{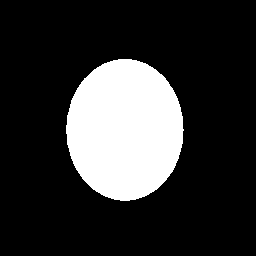}
\hspace{0.5mm}
\includegraphics[width=28mm]{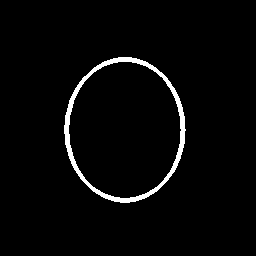}
\hspace{0.5mm}
\includegraphics[width=28mm]{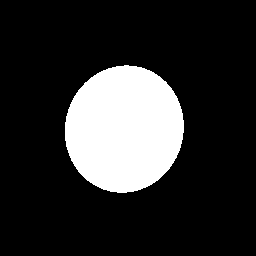}
\hspace{0.5mm}
\includegraphics[width=28mm]{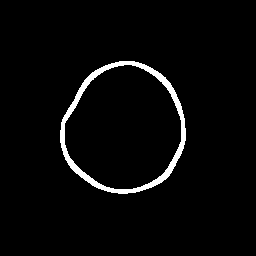}
\vspace{2mm}

\caption{From left to right: Image, Ground truth mask, Ground truth contour, Predicted mask, Predicted contour}
\label{fig:results2}
\end{figure}

\begin{figure}
\includegraphics[width=28mm]{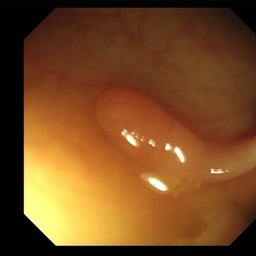}
\hspace{0.5mm}
\includegraphics[width=28mm]{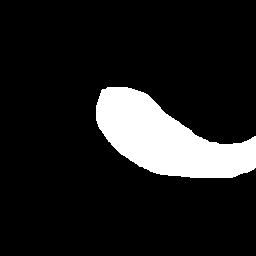}
\hspace{0.5mm}
\includegraphics[width=28mm,height=28mm]{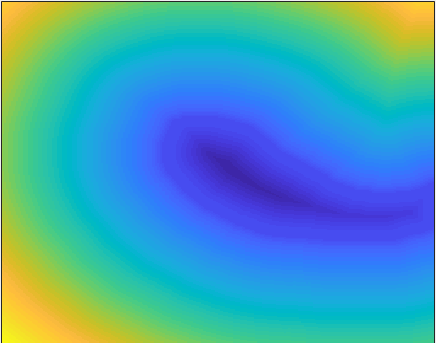}
\hspace{0.5mm}
\includegraphics[width=28mm]{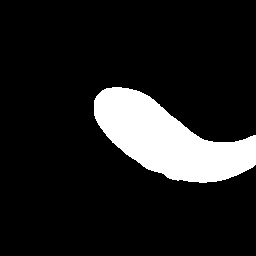}
\hspace{0.5mm}
\includegraphics[width=28mm,height=28mm]{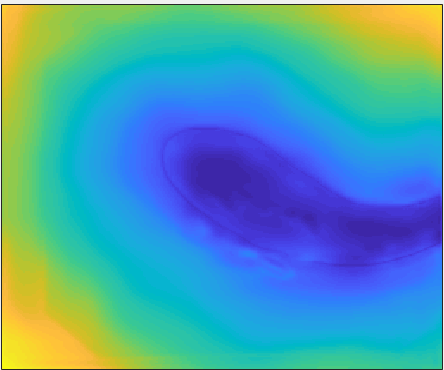}
\vspace{2mm}

\includegraphics[width=28mm]{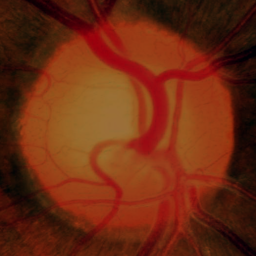}
\hspace{0.5mm}
\includegraphics[width=28mm]{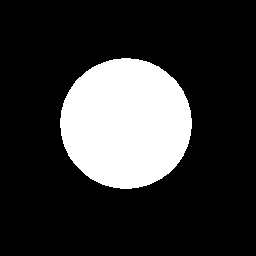}
\hspace{0.5mm}
\includegraphics[width=28mm,height=28mm]{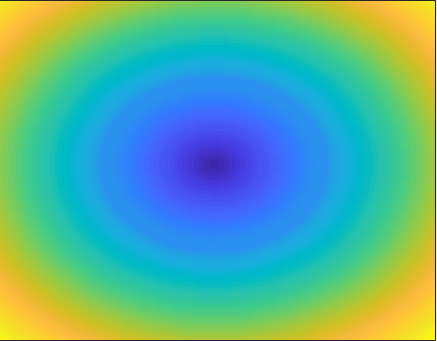}
\hspace{0.5mm}
\includegraphics[width=28mm]{images/GT_646_R.png}
\hspace{0.5mm}
\includegraphics[width=28mm,height=28mm]{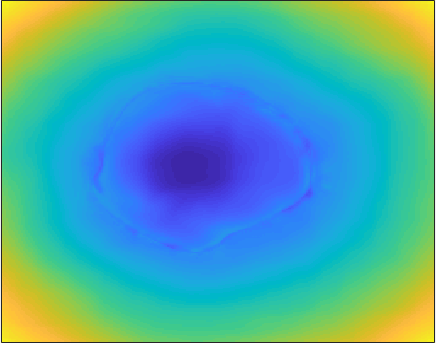}
\vspace{2mm}

\caption{From left to right : Image, Ground truth mask, Ground truth distance map, Predicted mask, Predicted distance map}
\label{fig:results3}
\end{figure}

The networks 1Enc 1Dec + Conv MC and 1Enc 1Dec + Conv MD outputs contour and distance along with the masks. This contour and distance maps helps in regulating the segmentation results.
In Figures \ref{fig:results2} and \ref{fig:results3} the predicted masks, contour and distance maps obtained are compared with the ground truth masks, contour and distance maps. From Figure \ref{fig:results2}, it can be seen that the predicted masks are the region filled versions of the estimated contours. A similar effect can also be seen in Figure \ref{fig:results3} where masks are contained by the predicted distance maps. And since, the distance map is obtained by regression we did not get a pixel level accurate  map but instead we get a map very close to the ground truth distance map. This shows how well distance maps are acting as regularizers.   

% Contour results polyp and retina

% Dist results polyp and retina

The difficulty of having accurate segmentation is attributed to variability in shape, texture, size, and color. Taking shape into consideration, polyp has higher variability when compared to cup and disc. Similarly, cup has higher variability when compared to disc. Because of this, disc has highest dice and jaccard when compared to both cup and polyp. This can be verified in Table \ref{table:metrics1}. %In this paper, we are using shape constraint to provide smooth and accurate segmentation.

\begin{figure}
\centering    
\includegraphics[width=28mm]{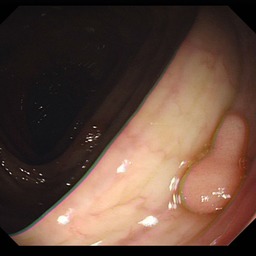}
\hspace{0.5mm}
\includegraphics[width=28mm]{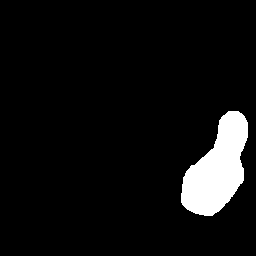}
\hspace{0.5mm}
\includegraphics[width=28mm]{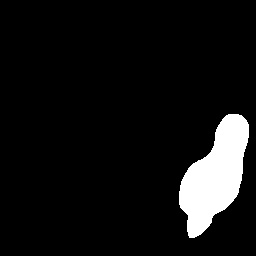}
\hspace{0.5mm}
\includegraphics[width=28mm]{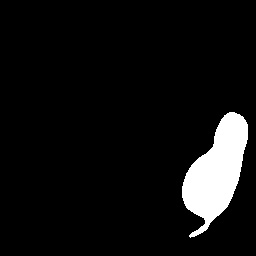}
\hspace{0.5mm}
\includegraphics[width=28mm]{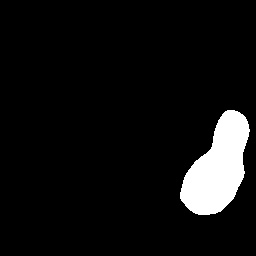}
\vspace{2mm}

\includegraphics[width=28mm]{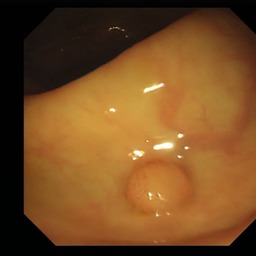}
\hspace{0.5mm}
\includegraphics[width=28mm]{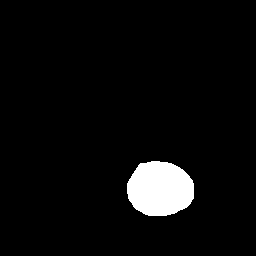}
\hspace{0.5mm}
\includegraphics[width=28mm]{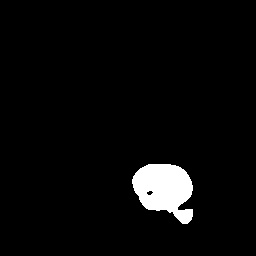}
\hspace{0.5mm}
\includegraphics[width=28mm]{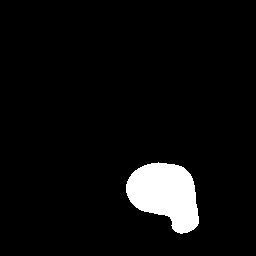}
\hspace{0.5mm}
\includegraphics[width=28mm]{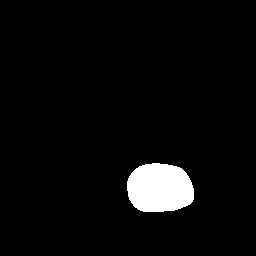}
\vspace{2mm}

\caption{From left to right : Image, Ground truth, Predicted mask with D1 as constraint, Predicted mask with D2 as constraint, Predicted mask with D3 as constraint.}
\label{fig:results4}

\end{figure}

So, in order to evaluate the effect of distance map, polyp and cup segmentation are the better choices. In Table \ref{table:metrics3}, the results of using D1, D2 and D3 distance maps as constraints for cup, disc, and polyp are shown. It can be seen that for disc there is not much difference in the scores. While for cup there is a slight improvement in using distance D3 over others. But for the case of polyp, using distance D3 shows considerable improvement over others.  In Figure \ref{fig:results4}, results obtained using three distance maps as regularizers are shown and compared with the ground truth. It is clear that the mask obtained by having distance D3 as a regularizer, gives a smooth and accurate segmentation compared to others. 

An intuitive explanation for this observation could be that the $l2-norm$ performs better in the absence of discontinuities and in the presence of smooth variations, and as seen in Figure \ref{fig:dist_maps}, the distance map D3 is smoother when compared to the distance maps D1 and D2 and hence could be the reason for its superior performance.

%is better suited for data withousuitable for sparse data, distance maps D1 and D2, whose representations are sparse when compared to D3, might have resulted in a lower performance.  

% This can be attributed to the usage of l2-norm which does'nt work well for sparse data. The distance maps D1, D2 have a sparse representation when compared to D3.   

\begin{table}[]
\centering
\caption{Comparison of segmentation metrics for different distance maps}
\label{table:metrics3}
\begin{tabular}{|c|c|c|c|c|c|c|}
\hline
\multirow{2}{*}{Distance map} & \multicolumn{2}{c|}{Cup} & \multicolumn{2}{c|}{Disc} & \multicolumn{2}{c|}{Polyp} \\ \cline{2-7} 
 & Dice & Jaccard & Dice & Jaccard & Dice & Jaccard \\ \hline
D1 & 0.8727 & 0.7812 & 0.9667 & 0.9362 & 0.7891 & 0.7020 \\ \hline
D2 & 0.8704 & 0.7779 & 0.9667 & 0.9362 & 0.8222 & 0.7396 \\ \hline
D3 & \textbf{0.8736} & \textbf{0.7831} & 0.9666 & 0.9361 & \textbf{0.8291} & \textbf{0.7514} \\ \hline
\end{tabular}
\end{table}

\section{Conclusion}\label{sect:conclusion}
In this paper, we proposed a deep multi-task network for the joint task of segmentation and shape learning. The network was shown to perform comparable to and in certain cases better than the previously proposed state-of-the-art FCNs, with an advantage of having lesser number of parameters and thereby consuming lesser time for training and inference. We also explored different ways in which spatial information can be incorporated and showed the impact of different distance maps on the segmentation tasks. A good future work would be to explore different ways of learning shape information other than contour map or distance map.         
% \midlacknowledgments{We thank a bunch of people.}

\bibliography{midl-samplepaper}
\newpage
\appendix
\section{Evaluation metrics}\label{appendix:A}
Jaccard index (also known as intersection over union, IoU) is defined as the size of the intersection divided by the size of the union of the sample sets, and it is calculated as follows:
\begin{equation}
    Jaccard(A,B) = \frac{|A \cap B|}{|A \cup B|}
\end{equation}
where A corresponds to the output of the method and B to the actual ground truth. 
\\
\\
DICE similarity score is a statistic also used for comparing the similarity of two samples. It is calculated as follows:
\begin{equation}
    Dice(X,Y) = \frac{2|X \cap Y|}{|X| + |Y|}
\end{equation}
where X and Y correspond, respectively, to the output of the method and the ground truth image. 
\newpage
\section{Feature map visualization}\label{appendix:B}

\begin{figure}[ht]
\centering    
\includegraphics[width=28mm]{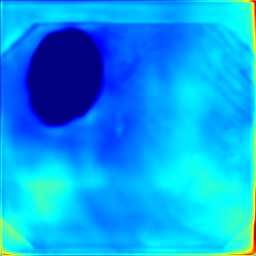}
\hspace{0.5mm}
\includegraphics[width=28mm]{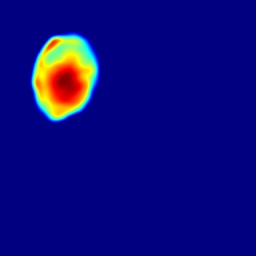}
\hspace{0.5mm}
\includegraphics[width=28mm]{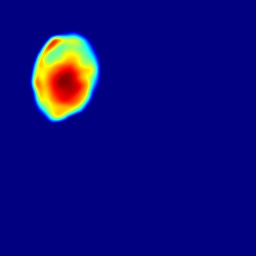}
\hspace{0.5mm}
\includegraphics[width=28mm]{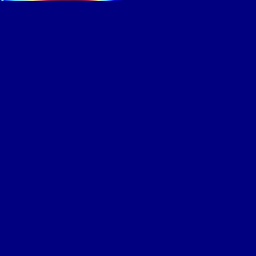}
\hspace{0.5mm}
\includegraphics[width=28mm]{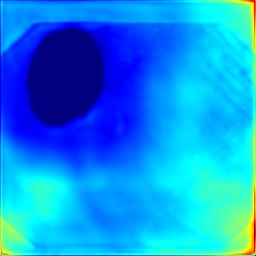}
\vspace{2mm}

\includegraphics[width=28mm]{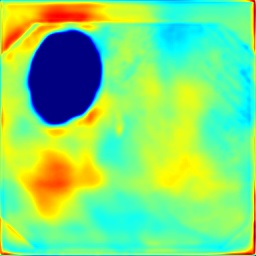}
\hspace{0.5mm}
\includegraphics[width=28mm]{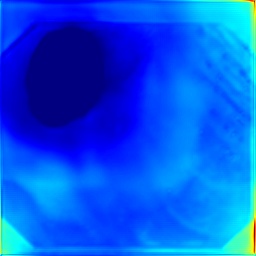}
\hspace{0.5mm}
\includegraphics[width=28mm]{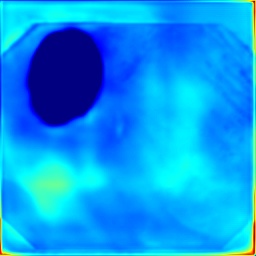}
\hspace{0.5mm}
\includegraphics[width=28mm]{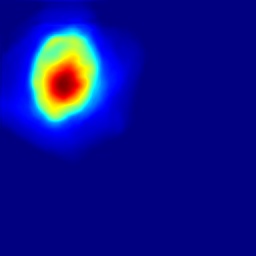}
\hspace{0.5mm}
\includegraphics[width=28mm]{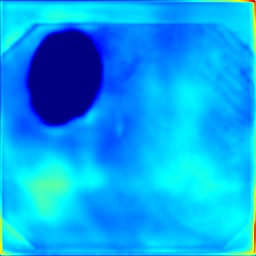}
\vspace{2mm}

\includegraphics[width=28mm]{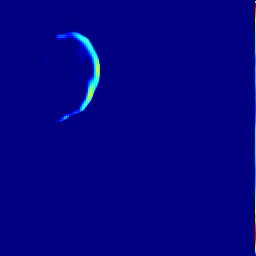}
\hspace{0.5mm}
\includegraphics[width=28mm]{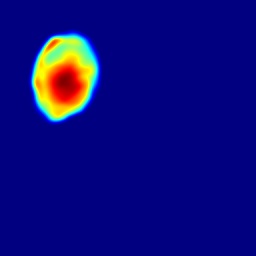}
\hspace{0.5mm}
\includegraphics[width=28mm]{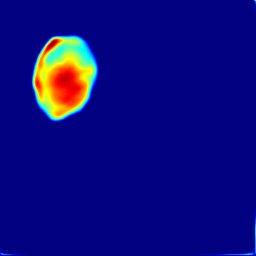}
\hspace{0.5mm}
\includegraphics[width=28mm]{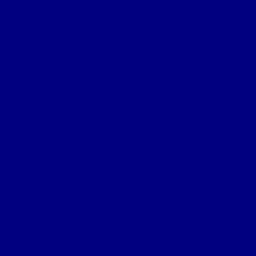}
\hspace{0.5mm}
\includegraphics[width=28mm]{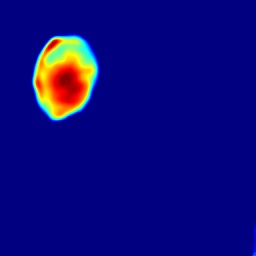}
\vspace{2mm}

\includegraphics[width=28mm]{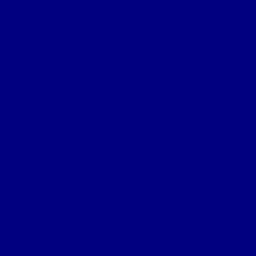}
\hspace{0.5mm}
\includegraphics[width=28mm]{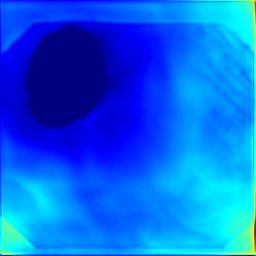}
\hspace{0.5mm}
\includegraphics[width=28mm]{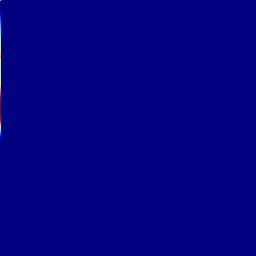}
\hspace{0.5mm}
\includegraphics[width=28mm]{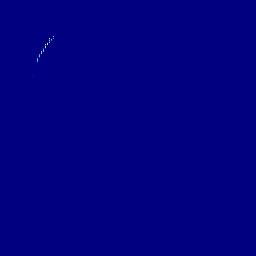}
\hspace{0.5mm}
\includegraphics[width=28mm]{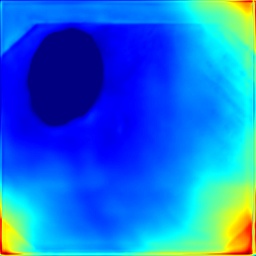}
\vspace{2mm}

\includegraphics[width=28mm]{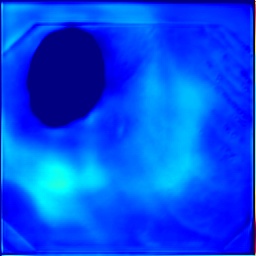}
\hspace{0.5mm}
\includegraphics[width=28mm]{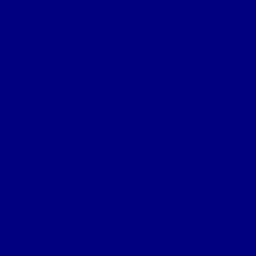}
\hspace{0.5mm}
\includegraphics[width=28mm]{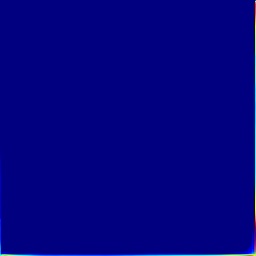}
\hspace{0.5mm}
\includegraphics[width=28mm]{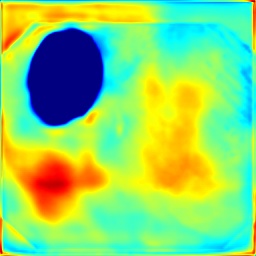}
\hspace{0.5mm}
\includegraphics[width=28mm]{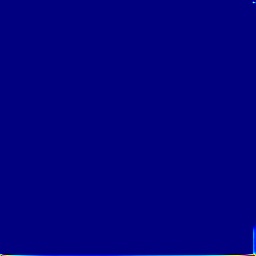}
\vspace{2mm}

\includegraphics[width=28mm]{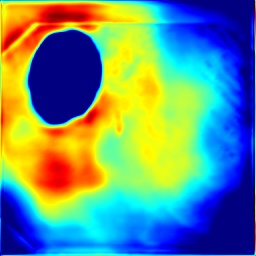}
\hspace{0.5mm}
\includegraphics[width=28mm]{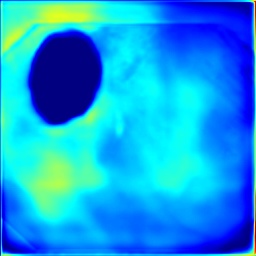}
\hspace{0.5mm}
\includegraphics[width=28mm]{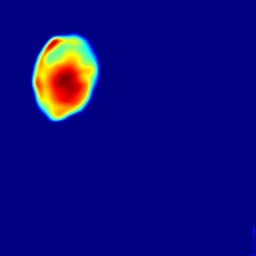}
\hspace{0.5mm}
\includegraphics[width=28mm]{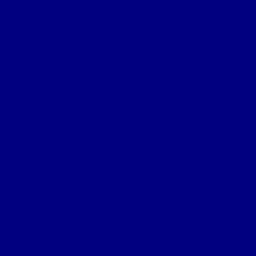}
\hspace{0.5mm}
\includegraphics[width=28mm]{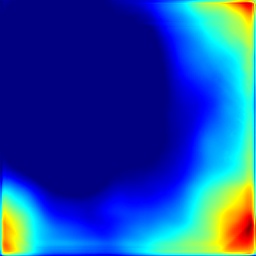}
\vspace{2mm}

\caption{Visualization of feature maps}
\label{fig:feature_map_vis}

\end{figure}

\end{document}